\documentclass[letterpaper, 10 pt, conference]{ieeeconf}
\IEEEoverridecommandlockouts                              % This command is only needed if
\usepackage{graphicx} % for pdf, bitmapped graphics files
\usepackage{epsfig} % for postscript graphics files
\usepackage{mathptmx} % assumes new font selection scheme installed
\usepackage{times} % assumes new font selection scheme installed
\usepackage{amsmath} % assumes amsmath package installed
\usepackage{amssymb}  % assumes amsmath package installed
\usepackage[mathscr]{eucal}
\usepackage{caption}
\usepackage{subcaption}
\usepackage{mathtools}
\usepackage{amsfonts} % |
\usepackage[ruled,vlined]{algorithm2e}
\usepackage{url}
\usepackage{multirow}
\usepackage{hyperref} % for autoref
\usepackage[table,xcdraw]{xcolor} % param table
\usepackage{placeins}
\usepackage{dsfont} % indicator function

\newcommand{\etal}{{\em et al.}}

\title{\LARGE \bf
Adaptive Curriculum Generation from Demonstrations for Sim-to-Real Visuomotor Control}

\author{Lukas Hermann$^*$, Max Argus$^*$, Andreas Eitel, Artemij Amiranashvili, Wolfram Burgard, Thomas Brox
\thanks{$^*$ First two authors contributed equally. All authors are with the University of Freiburg, Germany. Wolfram Burgard is also with the Toyota Research Institute, USA. This  work  has  been  supported  partly  by the BMBF grant No. 01IS18040B-OML and the DFG grant No. BR 3815/10-1.
{\tt\small \{hermannl,argusm,eitel\}@cs.uni-freiburg.de }}%
}

\begin{document}

\maketitle

%%%%%%%%%%%%%%%%%%%%%%%%%%%%%%%%%%%%%%%%%%%%%%%%%%%%%%%%%%%%%%%%%%%%%%%%%%%%%%%%
\begin{abstract}
We propose Adaptive Curriculum Generation from Demonstrations (ACGD) for reinforcement learning in the presence of sparse rewards. Rather than designing shaped reward functions, ACGD adaptively sets the appropriate task difficulty for the learner by controlling where to sample from the demonstration trajectories and which set of simulation parameters to use. We show that training vision-based control policies in simulation while gradually increasing the difficulty of the task via ACGD improves the policy transfer to the real world. The degree of domain randomization is also gradually increased through the task difficulty. We demonstrate zero-shot transfer for two real-world manipulation tasks:
%not solvable by state-of-the-art deep reinforcement learning methods: 
pick-and-stow and block stacking.
A video showing the results can be found at \url{https://lmb.informatik.uni-freiburg.de/projects/curriculum/}
%- We consider the problem of reinforcement learning in the presence of sparse reward
%- visuomotor policy, solve robotic manipulation tasks in the real world.
%- sim2real transfer
% two real-world tasks, block-stacking and box-placement
\end{abstract}

%%%%%%%%%%%%%%%%%%%%%%%%%%%%%%%%%%%%%%%%%%%%%%%%%%%%%%%%%%%%%%%%%%%%%%%%%%%%%%%%
\section{INTRODUCTION}

Reinforcement Learning (RL) holds the promise of solving a large variety of manipulation tasks with less engineering and integration effort. Learning continuous visuomotor controllers from raw images circumvents the need for manually designing multi-stage manipulation and computer vision pipelines~\cite{wirnshofer18}. Vision-based closed-loop control can increase robustness of robots performing real-world fine manipulation tasks. Prior experience has shown that learning-based methods can cope better with complex semi-structured environments that are hard to model in a precise manner due to contacts, non-rigid objects or cluttered scenes~\cite{8461044,zeng2018learning,eitel17isrr}.    
%These solutions should be more scaleable for complex semi-sructured environments with clutter and soft objects.

A major challenge preventing deep reinforcement learning methods from being used more widely on physical robots is exploration. RL algorithms often rely on random exploration to search for rewards. 
This hinders the application to real-world robotic tasks in which the reward is too sparse to ever be encountered through random actions. Additionally, in the real world random exploration can also be dangerous for the robot or its environment.
%In many cases specifying sparse rewards is more desirable as a desired end state is more straightforward to evaluate.

A common strategy to address the exploration problem is reward shaping, in which distance measures and intermediate rewards continuously provide hints on how to reach the goal~\cite{Ng99policyinvariance}. Reward shaping is typically task specific and requires manual setup as well as careful optimization.
This contradicts the purpose of using RL as a general approach and can bias policies to satisfy shaped rewards.

Poor exploration can also be compensated by providing demonstration trajectories and performing Behavior Cloning (BC)~\cite{zhang2018deep}.
Pre-training with behavior cloning can guide the RL algorithm to initial rewards, from where it can be optimized further~\cite{Rajeswaran-RSS-18}. Behavior cloning usually requires a substantial amount of expert demonstrations.

\begin{figure}[t]
    \centering
    \includegraphics[width=0.99\columnwidth]{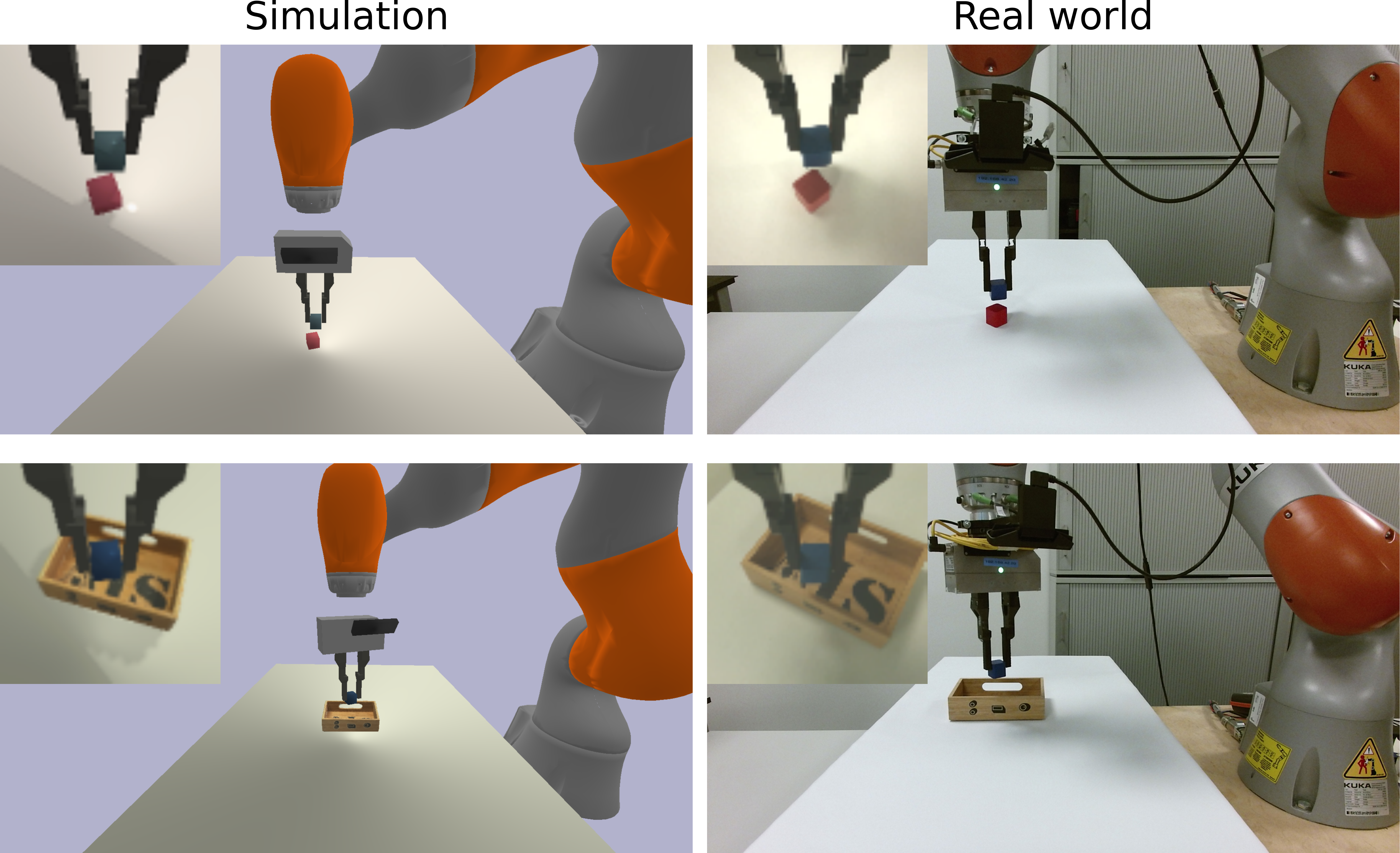}
    \caption{Adaptive Curriculum Generation from Demonstrations utilizes only 10 demonstration trajectories to enable learning visuomotor policies for two fine manipulation tasks: block stacking (top) and pick-and-stow (bottom). The visuomotor policies are trained in simulation and transferred without further training to a physical robot. The image-in-image shows the egocentric view of the arm-mounted RGB camera. The third person view is not provided to the robot.}
    \label{fig:coverfigure}
\end{figure}
% max: what are these, they prevent me from using negative vpsaces...
%\thispagestyle{empty}
%\pagestyle{empty}
This paper tackles the exploration problem with curriculum learning based on a few manual demonstrations. We present an adaptive curriculum generation algorithm that controls difficulty by controlling how initial states are sampled from demonstration trajectories as well as controlling the degree of domain randomization that is applied during  training.
The algorithm continually adapts difficulty parameters during training to keep the rewards within a predetermined, desired success rate interval.
Consequently, the robot can always learn at the appropriate difficulty level, which speeds up the convergence of the training process and the final generalization performance. The method is simple and applicable to a variety of robotic tasks. 
We demonstrate its performance on real world pick-and-stow and block stacking tasks. 
%As we still struggle to find ways to autonomously acquire sufficient amounts of training data in the real world with the robot, 
We apply the curriculum learning in conjunction with policy learning in a physics simulator. Afterwards, the learned policies are transferred without further re-training to the physical robot.
The advantages of training in simulation include drastically increased learning speed through distributed training, no human involvement during training, improved safety, and added algorithmic possibilities due to  access to simulator state information. 

We present a novel perspective on domain randomization inside our curriculum learning framework. Our algorithm automatically learns when to increase the amount of visual domain randomization and dynamics randomization during the training process, until the policy exhibits the desired degree of domain invariance required for a successful simulation to reality transfer.

The main contributions of this paper are:  1) a curriculum generation method for learning with sparse rewards that only requires a dozen demonstrations, 2) an algorithm for automatic and controlled scaling of task difficulty, 3) a unified treatment of demonstration sampling and domain randomization as task difficulty parameters of the curriculum, 4) zero-shot transfer from simulation to real-world for two robot manipulation tasks.

%Contributions: \\
%1) Learning with sparse rewards with the use of very few demonstrations\\
%2) Algorithm with controlled scaling of task difficulty\\
%3) Unified treatment of demonstration sampling and domain randomization as difficulty parameters\\

%1) unified treatment of demonstrations and domain randomization \\
%5) Train several tasks with the same parameters \\
%2) sampling along trajectory, which allows more complicated tasks than sampling end states. \\
%3) Requires very few demonstrations \\
%4) Transfer vision-based policy to real-world. \\
%5) Not like imitation, don't follow actions.

%What curriculum has to do with simulation. From scratch learning in real-world hard, safety concerns, long training times, high material costs, calculating reward signal requires exactly the manual engineering which you try to avoid with RL, therefore learning in simulation easier.
%In simulation we can control state and parameters of environment, good for curriculum learning, always learn at appropriate difficult level.

\section{RELATED WORK}

\subsection{Training Curricula}
%Our method builds on prior research that aims for faster training of machine learning methods by learning with gradually increasing difficulty. 
The seminal paper by Bengio~\etal~\cite{Bengio:2009} shows that curriculum learning provides benefits for classical supervised learning problems such as shape classification.
Florensa~\etal~\cite{florensa2017reverse} propose a reverse curriculum for reinforcement learning that requires only the final state in which the task is achieved.
Their curriculum is generated gradually by sampling random actions to move further away from the given goal and thus reversely expanding the start state distribution with increasingly more difficult states. 
The key difference of our method is that we extend the idea of curriculum learning to task difficulty parameters, which goes beyond sampling start states nearby the goal. 
Further, we sample backwards from the demonstration trajectory in comparison to randomly sampling backwards in action space, which is likely to fail for difficult states that are unlikely to be encountered randomly.
%We show that controlling the task difficulty and the amount of domain randomization during training by means of a curriculum is crucial for generalization performance. 
Another approach is to generate a linear reverse curriculum, a baseline that we compare against in our experiments~\cite{fan2018surreal}. A more evolved approach generates a curriculum of goals using a Generative Adversarial Network (GAN) but cannot handle visual goals~\cite{pmlr-v80-florensa18a}.
Learning by playing also generates a form of curriculum, where auxiliary tasks are leveraged to learn increasingly more complicated tasks~\cite{DBLP:journals/corr/abs-1802-10567}. Nevertheless, it is not straightforward to design auxiliary tasks that can benefit the final goal task.

\subsection{Learning from Demonstrations and Imitation}
The most common approach for learning from demonstrations is supervised learning, also known as behavior cloning. Behavior cloning has shown to enable training of successful manipulation policies~\cite{zhang2018deep}, but usually has to cope with the compounding-error problem and requires hundreds or even thousands of expert demonstrations. Several methods have combined behavior cloning with RL, either in a hierarchical manner ~\cite{strudel2019combining} or as a pre-training step to initialize the RL agent~\cite{Rajeswaran-RSS-18}. Other methods leverage demonstration data for training off-policy algorithms by adding demonstration transitions to the replay buffer~\cite{vevcerik2017leveraging}. In comparison to the those works our method does not try to copy the demonstrated policy. Only the visited states are used as initial conditions and the actions taken during demonstrations are not required.
% require to store the actions taken during demonstrations but only the visited states.
Therefore, our method is more robust against suboptimal demonstrations. 

Demonstration data  can  also be  provided in form of raw video data~\cite{sermanet2018time, aytar2018playing,tai18, mees2019adversarial}.
These approaches can work if sufficient demonstration  data  is  provided, while our method requires only around a dozen demonstrations. Nair~\etal~\cite{nair2018overcoming} combine off-policy learning with demonstrations to solve block stacking tasks. They use Hindsight Experience Replay~\cite{andrychowicz2017hindsight} for multi-goal tasks, but this method is not applicable to visual domains where the desired goal configuration is not given, as in our tasks. Zhu~\etal~\cite{Zhu-RSS-18} combine imitation learning and RL for training manipulation policies in simulation. They report preliminary success for transferring the learned policies into the real world but also challenges, due to different physical properties of simulation and the real world. We address this problem by leveraging our curriculum generation approach for gradually training with more realistic simulation parameters.

\subsection{Sim-to-Real and Domain Randomization}
For sim-to-real transfer of our policy we build upon an efficient technique called domain randomization~\cite{tobin2017domain}. It has been successfully applied to transfer imitation learning policies for a pick-and-place task on a real robot~\cite{pmlr-v78-james17a}. 
Since our visuomotor controller operates in a first-person view with sensor data from a camera-in-hand robot setup, there is significantly less background clutter, so we generally use a smaller degree of domain randomization. 
Nevertheless, as we show in experiments domain randomization can harm convergence during RL training, so we circumvent this problem by incorporating domain randomization into our adaptive curriculum.
Domain randomization has also been extended to dynamics randomization~\cite{peng2018sim}, which we also incorporate into our approach. A recent method proposes to close the sim-to-real gap by adapting domain randomization with experience collected in the real world~\cite{Chebotaricra19}. 
Experiments with a real robot show impressive results for a drawer-opening and a peg-in-hole task, but the learned policies are not based on visual input and it is unclear if the method also works with sparse rewards. 
In concurrent work, OpenAI \etal~\cite{openai2019rubiks} have also considered adaptive domain randomization for sim-to-real dexterous manipulation.

Sim-to-real transfer can also be viewed from a transfer learning or domain adaptation perspective~\cite{pmlr-v78-rusu17a}. Here, methods that use GANs have been successfully applied for instance grasping on a real robot~\cite{bousmalis2018using}. Nevertheless, training the GAN requires a great amount of real-world training data (on the order of 100,000 grasps). Recent meta-learning methods seem to be suitable to reduce the amount of data needed in the real world for efficient domain adaptation~\cite{Yu-RSS-18}. %Transfer learning to the real world can also be performed using progressive networks. 
Those transfer learning methods are outside of the scope of our work because we focus on zero-shot sim-to-real transfer.

%Related work motor control
\subsection{Motor Control for Manipulation }
Model-based reinforcement learning techniques with probabilistic dynamics models have been proposed to learn control policies for block stacking~\cite{Deisenroth-RSS-11} and multi-phase manipulation tasks~\cite{kroemer2015towards}. Guided Policy Search has been applied to visuomotor control tasks~\cite{JMLR:v17:15-522}. The mentioned approaches work well in the local range of trained trajectories, but generalize less to larger variations in goal and robot state positions, which is required for our tasks. %A recent method uses model-free reinforcement learning to train visuomotor policies only in the real world to grasp objects in bins, but requires very large amounts of real training data~\cite{kalashnikov18a}. 
%We use a model-free deep reinforcement learning algorithm for visuomotor control to generalize across variations of object positions and we train only in simulation.

\section{ADAPTIVE CURRICULUM GENERATION FROM DEMONSTRATIONS}
%This section has to be independent 

%> We want to simplify the task to make it easy to solve and then gradually increase the difficulty over time
%> introduction of the parameter: scaled by the success rate
%> list of all difficulty settings
%> Demonstration sampling:
%> >   Don't want to do reward shaping with distances, we want to use sparse rewards
%> Domain randomization
%> Task setting:
%> > Physik
%> > initial conditions
%> explain why 2 difficulty parameter, (works better empirically)

% The adaptive curriculum generation method we propose controls the simulated environment to make the task easier to solve and then gradually increases the difficulty over time.
% This is done by creating a difficulty parameter influencing the difficulty of the task and scaling this based on the success rate of solving the task at the current difficulty level.
% The difficulty parameter controls difficulty by determining how initial states from demonstration trajectories are sampled and controlling the degree of domain randomization.
% Our method combines learning from demonstrations with curriculum generation in order to enable learning complex tasks from sparse rewards. In the following sections we describe each of these components and how they are combined.
In this section we present Adaptive Curriculum Generation from Demonstrations (ACGD), a method to overcome exploration difficulties of reinforcement learning from sparse rewards in simulated environments.
Depending on the current success rate, ACGD automatically schedules increasingly difficult subtasks by shaping the initial state distribution and scaling a set of parameters that control the difficulty of the environment, such as the degree of domain randomization.
% Requiring only a handful of manual demonstrations, it generates a reverse curriculum of start states for the RL agent.
% Compared to previous approaches with uniform or linear curricula, our approach exploits task properties by adaptively regulating the difficulty of subtasks.  

\subsection{Reinforcement Learning}
%\subsection{}
%Reinforcement learning introduction, sample inefficiency

In reinforcement learning an agent makes some observation ($o_t$) of an underlying environment state, which is used by a policy to compute an action $a_t = \pi(o_t)$. This produces transitions consisting of $(o_t,a,r,o_{t+1})$ for discrete timesteps. In a sparse reward setting a correct sequence of actions produces rewards ($r_t$). The policy is optimized to maximize the discounted future rewards, $R=\sum_{i=t}^T \gamma^{(i-t)} r_i$, called return.

A number of different RL algorithms exist, they can be categorized as either off-policy algorithms or on-policy algorithms based on if they make use of transitions that are not generated by the current policy being optimized. In our work we use the Proximal Policy Optimization (PPO)~\cite{DBLP:journals/corr/SchulmanWDRK17} algorithm, as on-policy algorithms are more suited to gradually changing the environment parameters.

\subsection{Reverse-Trajectory Sampling}
During training in simulation it is possible to initialize episodes from arbitrary states of demonstration trajectories.
We bootstrap exploration by using reverse-trajectory sampling, meaning we initially sample states from the end of the demonstration trajectories. These states are close to achieving sparse rewards, making these tasks much easier and solvable using RL. As the training progresses we successively sample states closer to the beginning of demonstration trajectories. This creates a curriculum in which the agent learns to solve a growing fraction of the task.

\begin{algorithm}
\SetAlgorithmName{Algorithm}{algorithm}{list of algorithms name}
\SetAlgoLined
\SetKwFunction{FMain}{train\_policy}
\SetKwInOut{Input}{Input}
\SetKwInOut{Output}{Output}
\SetKwProg{With}{with}{ do}{}
\SetKwProg{Other}{otherwise}{}{}
\Input{Iterations $N$, initial policy $\pi_0$, increment $\epsilon$, task params $\mathcal{H}^j=\{\mu_{init}^j, \sigma_{init}^j, \mu_{end}^j, \sigma_{end}^j\}$, reward interval $[\alpha, \beta]$ }
\Output{final policy $\pi_{N}$}    
\nl$sr_d,~sr_r,~\delta_d,~\delta_r \leftarrow 0$\;
\nl     \For{$i\leftarrow 1$ \KwTo $N$}{
\nl         \With{probability $p=0.5(sr_r + i/N)$}{
%\nl             regular restart\;
\nl             $\texttt{sample\_regular\_restart}(\mathcal{H},~\delta_r)$\;
\nl             $sr,~\delta \leftarrow sr_r,~\delta_r$\;
            }
\nl         \Other{} {
%\nl             demonstration restart\;
\nl             $\texttt{sample\_demonstration\_restart}(\mathcal{H}_d,\delta_d)$;
\nl             $sr,~\delta \leftarrow sr_d,~\delta_d$\;
           }
\nl        $ rollouts \leftarrow \texttt{generate\_rollouts}(\pi)$\;
\nl        $ \pi \leftarrow \texttt{update\_policy}(rollouts,~\pi)$\;
\nl        $ sr \leftarrow \texttt{update\_success\_rates}(rollouts)$\;
%\nl        $ \delta  \leftarrow \texttt{update\_difficulties}(\alpha, ~\beta, ~sr)$\;
\nl        $\delta \leftarrow \delta + \epsilon \cdot \mathds{1}_{(sr > \beta)} - \epsilon \cdot \mathds{1}_{(sr < \alpha)}$\;
}
%
% I think this is clearest without any indexes
%
\caption{Adaptive Curriculum Generation from Demonstrations}
\label{algo:acg}
\end{algorithm}

However, sampling states from the demonstration trajectories restricts the policy to observing initial states present in the recorded demonstrations. Especially if we want our algorithm to work with few demonstrations this presents a potential source of bias. To overcome this problem we mix resets from demonstrations with regular resets of the environments which have automatic randomization of initial states.
The choice of sampling demonstration or regular resets is made by a mixing function depending linearly on the success rates $sr_r$ of regular resets episodes and the overall progress of the training.
As opposed to \cite{fan2018surreal}, who claim that it is important for reverse curriculum learning to include previously sampled sections throughout the training in order to prevent catastrophic forgetting, our experiments suggest that this is not the case.

\subsection{Task and Environment Parameter Adaptation}
Apart from the distance between start states and goal states, the difficulty of a task also depends on a set of factors, such as the degree of domain randomization, intrinsics of the physics simulator or criteria that define the task completion.
As an example, the complexity of block stacking depends significantly on the bounciness of the blocks.
Therefore, we design our tasks such that their difficulty can be controlled by a set of parameters $\mathcal{H}$.
Examples can be found in Table~\ref{tab:task_params}, most of these are not task specific.
At the beginning of the training all parameters are set to the intuitively easiest configuration (e.g. less bouncy objects or smaller initial distance between objects).
The parameters that determine the degree of appearance or dynamics randomization are initially set to a minimum.
During training we scale the variance of the sampling distributions  and thus increase the difficulty by linearly interpolating between 0 and the maximal value chosen based on what is realistic.
Our experiments clearly suggest, that it is beneficial to gradually increase the degree of randomization over time since too much domain randomization from the beginning slows down training or might even prevent the policy from learning the task at all. 
To our knowledge, we are the first to apply curriculum learning to sim-to-real transfer.
When sampling initial states from demonstration it is not possible to randomize all parameters because some configurations are pre-determined by the demonstration. This results in a different set of parameters being randomized for each type of reset, see Table~\ref{tab:task_params}.

\begin{figure}[tb]
    \centering
    \includegraphics[width=0.95\columnwidth]{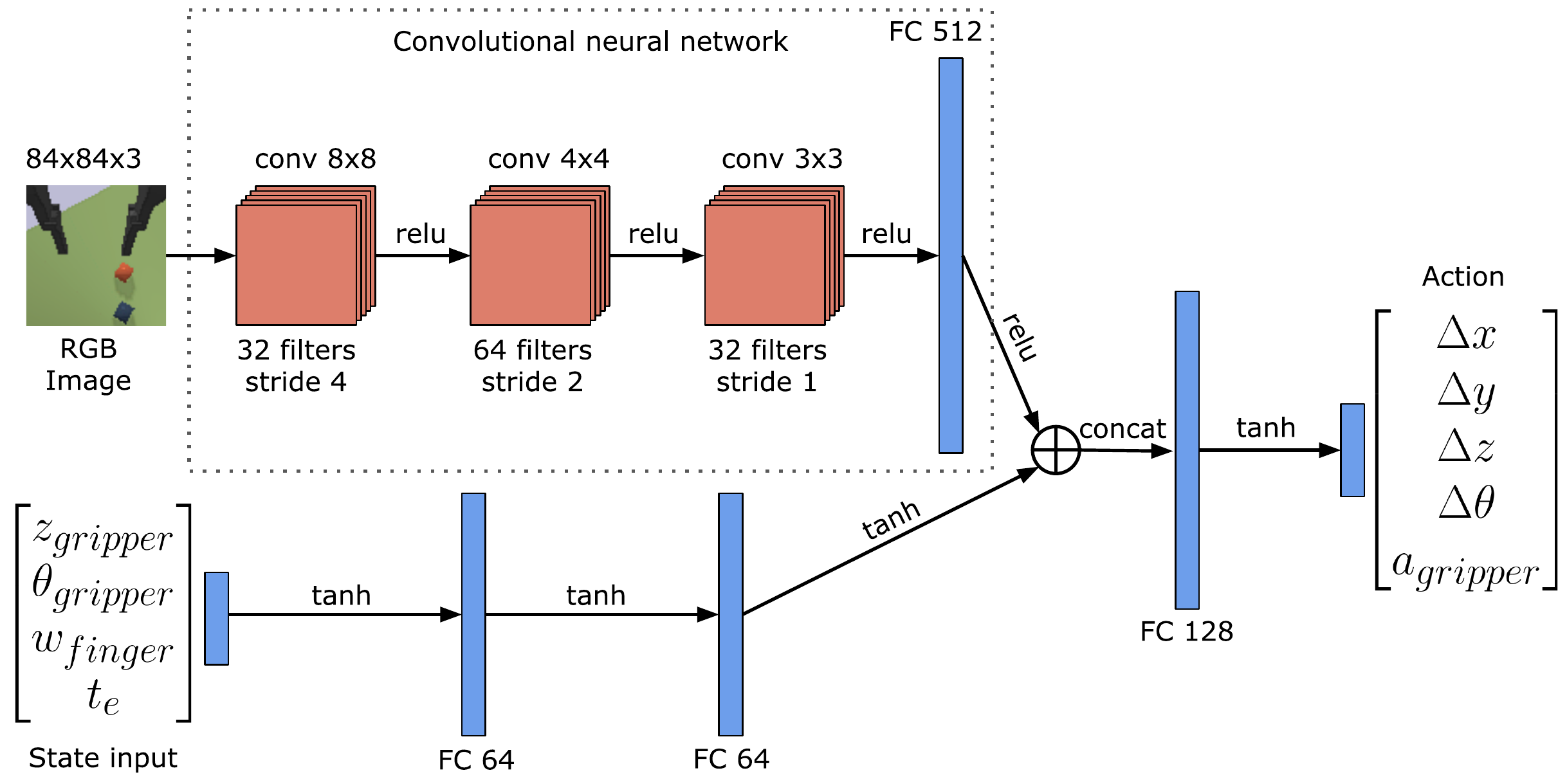}
    \caption{Architecture of the policy network. 
    The value function has the same architecture apart from having a single output for the value. 
    Policy and value functions share the weights of the CNN (dotted box).}
    \label{fig:network}
\end{figure}
%\vspace{-1em}

\subsection{Adaptive Curriculum Generation}
The challenge of curriculum learning is to decide a good strategy to choose the appropriate difficulty of start states and task parameters in the course of the training.
Previous approaches sampled initial states uniformly \cite{DBLP:journals/corr/PopovHLHBVLTER17}, \cite{nair2018overcoming}, \cite{Zhu-RSS-18} or linearly backwards \cite{fan2018surreal}.

However, sampling states from the end of the demonstration trajectories for too long unnecessarily slows down the training because the policy is trained on subtasks that it has already learned to master.
On the other hand, sampling more difficult task configurations too fast may prevent the policy from experiencing any reward at all.
Especially tasks with long episode lengths often do not have a constant difficulty at every stage.
Consider for instance a stacking task: it consists of both easier parts that only require straight locomotion and more difficult bottleneck moments like grasping and placing the object.
Our method adaptively generates a curriculum, such that the learning algorithm automatically dedicates more time on hard parts of the task, while not wasting time at straightforward sections.

Intuitively, we want the probability of experiencing a reward to be neither too high, nor too low.
Instead, our goal is to confine the probability within a desired reward region $\alpha \leq \mathbb{P}(R_t > 0 \vert \pi) \leq \beta$,
where $R_t$ denotes the return of a rollout started from a state sampled from the demonstration data at timestep $t$ and the interval $[\alpha, \beta]$ are hyperparameters which we set to $[0.4, 0.6]$
after empirical testing.
This is inspired by the \textit{Goals of Intermediate Difficulty} of \cite{pmlr-v80-florensa18a}.
For sparse bi-modal tasks, the probability $\mathbb{P}(R_t > 0 \vert \pi)$ corresponds to the expected success rate of the task. 

We control all difficulty parameters with two coefficients $\delta_d$ and $\delta_r \in [0, 1]$, which regulate the difficulty of resets from demonstrations and regular resets respectively by scaling the variances of $\mathcal{H}$ linearly w.r.t $\delta$.
A $\delta$ close to 0 corresponds to the easiest and close to 1 corresponds to the most difficult task setting (i.e. initializing episodes further away from the goal and sampling task parameters with higher variance).
Our method tunes the difficulty during training to ensure that the success rates of regular and demonstration resets ($sr_r$ and $sr_d$) stay in the desired region, as shown in the pseudo-code in Algorithm \ref{algo:acg}.
An example of this optimization can be seen in Fig. \ref{fig:comparison_stack}.

\section{EXPERIMENTS}

We demonstrate the effectiveness of the proposed curriculum generation via two manipulation tasks: pick-and-stow and stacking small blocks; see Fig.~\ref{fig:coverfigure}.
We posed the following questions: 1) can curriculum learning with demonstrations enable learning tasks with sparse rewards that do not succeed without curriculum learning?
2) what is the generalization performance compared to the existing behavioral cloning and reinforcement learning with shaped rewards? 
3) does our adaptive curriculum outperform other curriculum baselines?
4) can visuomotor policies trained in simulation generalize to a physical robot?

\subsection{Experimental Setup}
For training and evaluation of the policies we re-created the two tasks and the robot setup in a physics simulator. We aligned simulation and real-world as closely as possible; for a side-by-side comparison see Fig.~\ref{fig:coverfigure}.

Our KUKA iiwa manipulator is equipped with a WSG-50 two finger gripper and an Intel SR300 camera mounted on the gripper for an eye-in-hand view. The control of the end effector is limited to the rotation around the vertical z-axis, such that the gripper always faces downwards, it is parameterized as a reduced 5 DoF continuous action $a= [\Delta x, \Delta y, \Delta z, \Delta \theta, a_{gripper}]$ in the end effector frame.
Here, $\Delta x,y,z$ specify a Cartesian offset for the desired end effector position, $\Delta \theta$ defines the yaw rotation of the end effector and $a_{gripper}$ is the gripper action that is mapped to the binary command open/close fingers.

The observations that the policy receives are a combination of the $84 \times 84$ pixels RGB camera image and a proprioceptive state vector consisting of the gripper height above the table, the angle  that specifies the rotation of the gripper, the opening width  of the gripper fingers and the remaining timesteps of the episode normalized to the range  $[0,1]$, see Fig.~\ref{fig:network}.

Consistent with results of~\cite{47141}, preliminary experiments showed that it is beneficial to include the proprioceptive features.
The policy network is trained using PPO, but our approach can be used with any on-policy RL algorithm. 
For all experiments, training runs 8 environments in parallel for a total number of $10^7$ timesteps, this takes approximately 11 hours on a system with one Titan X and 16 CPU cores. During training the performance of the policy is always evaluated on the maximum task difficulty in order to obtain comparable results between different methods.
The policy output is the mean and standard deviation of a diagonal Gaussian distribution over the 5-dimensional continuous actions.
The value function yields a scalar value as output. We use the \textit{Adam}~\cite{DBLP:journals/corr/KingmaB14} optimizer with an initial learning rate of $0.00025$, which is linearly decayed in the course of the training.

%\argus{Refer to list of parameters that are controlled here? Put in appendix?}
%\lukas{we dont need appendix. but is there enough space for a full list of params?}
%\argus{I think we should hahve it, I'm putting stuff in the appendix for now to have it somewhere, we can merge the content into the main sections later.}

\begin{figure}[tb]
    \centering
    \includegraphics[width=1.0\columnwidth]{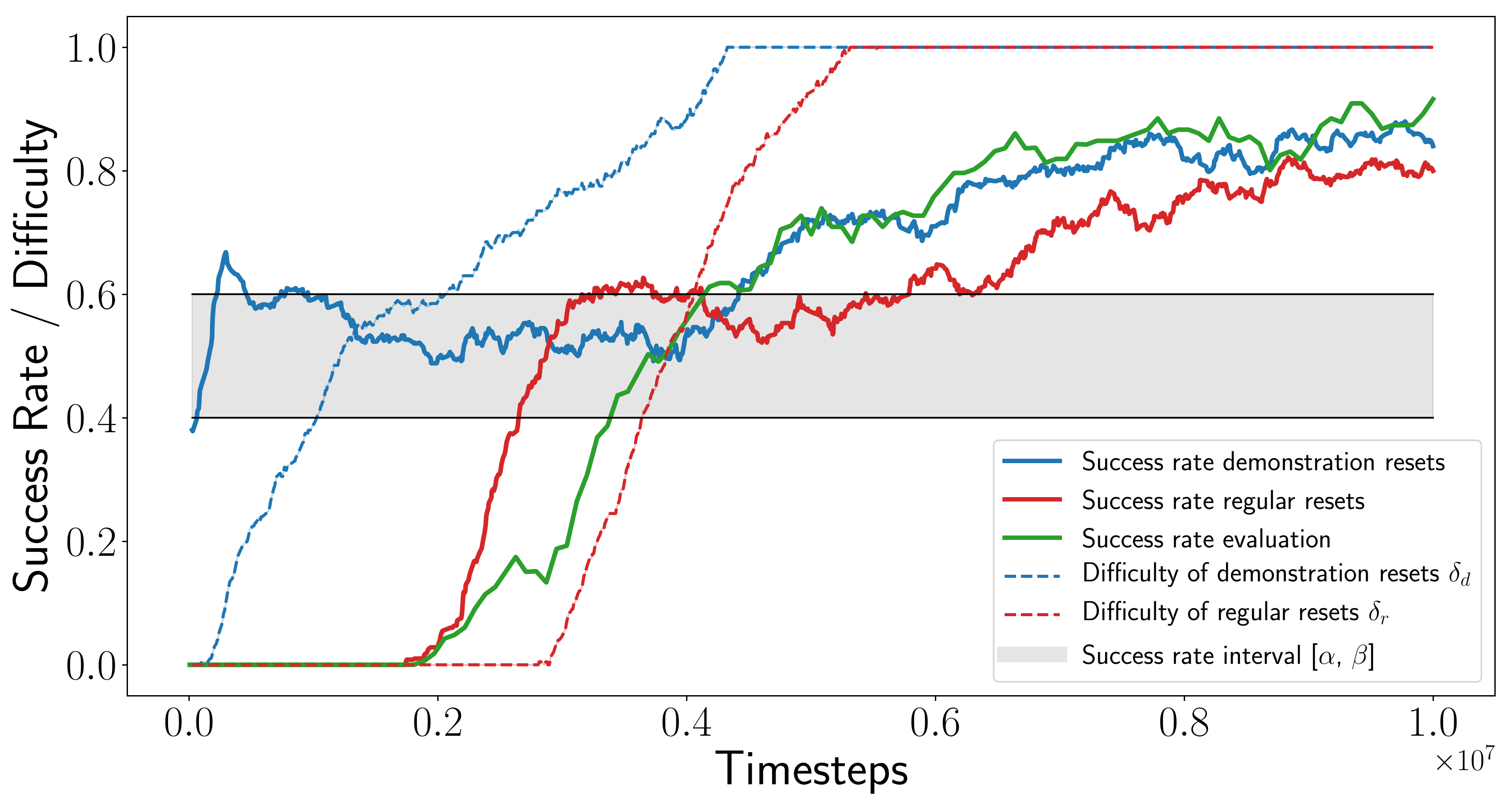}
    \caption{Example training run showing evolution of success rates and difficulty coefficients $\delta$ over the course of training. First the success rate and thus difficulties increase for resets from demonstrations (blue curves), followed by increases for regular resets (red). The plot shows that the success rate is kept in the desired interval (grey area) by the difficulty coefficients until the highest difficulty is reached. 
    %~\andreas{need to stress that this shows how adaptive method is}
    }
    \label{fig:comparison_stack}
\end{figure}

\begin{figure}[tb]
    \centering
    \includegraphics[width=1.0\columnwidth]{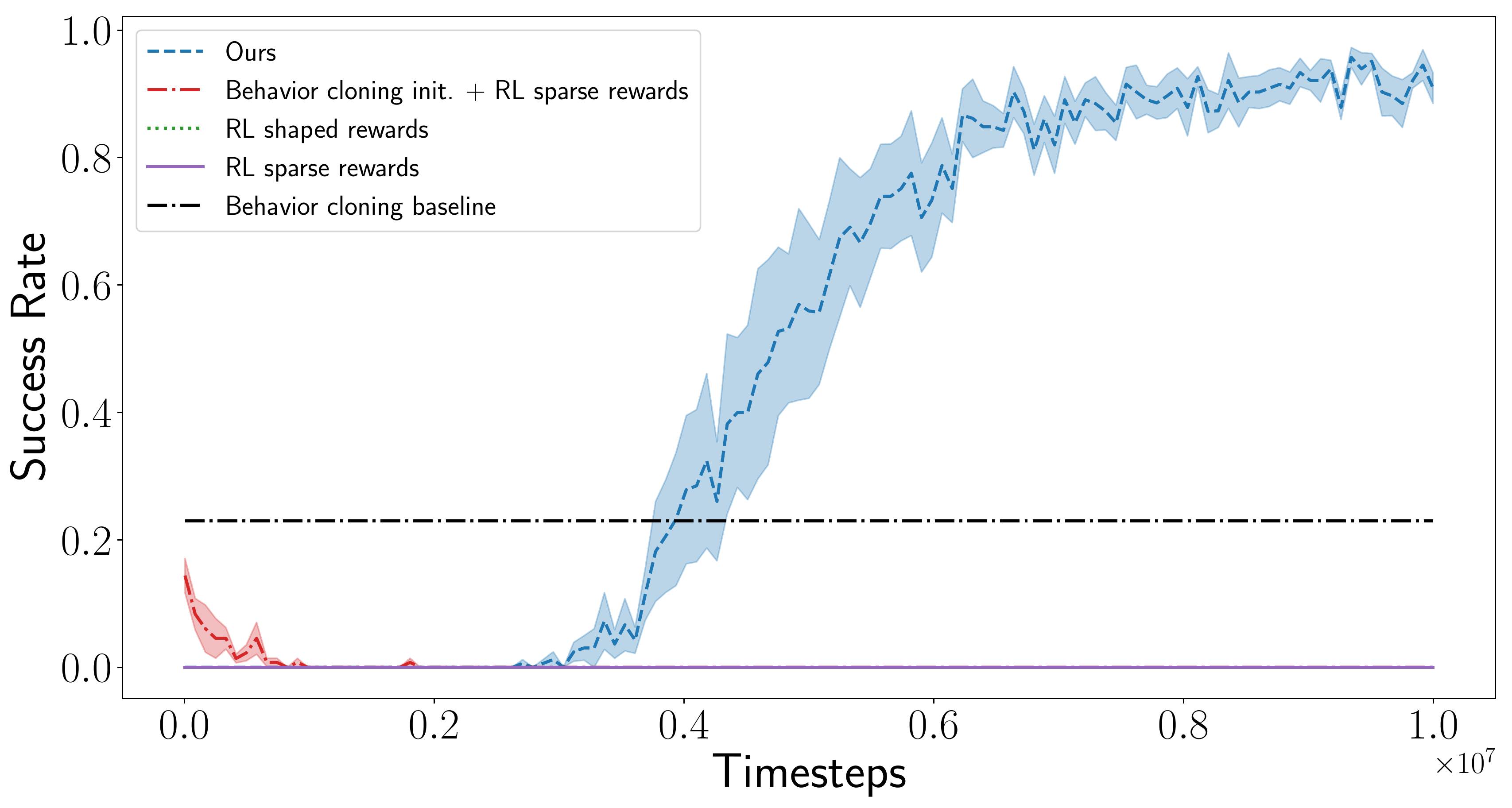}
    \caption{ACGD vs. baseline methods for the pick-and-stow task, evaluated in simulation.}
    \label{fig:comparison_box}
\end{figure}

\begin{figure}[tb]
    \centering
    \includegraphics[width=1.0\columnwidth]{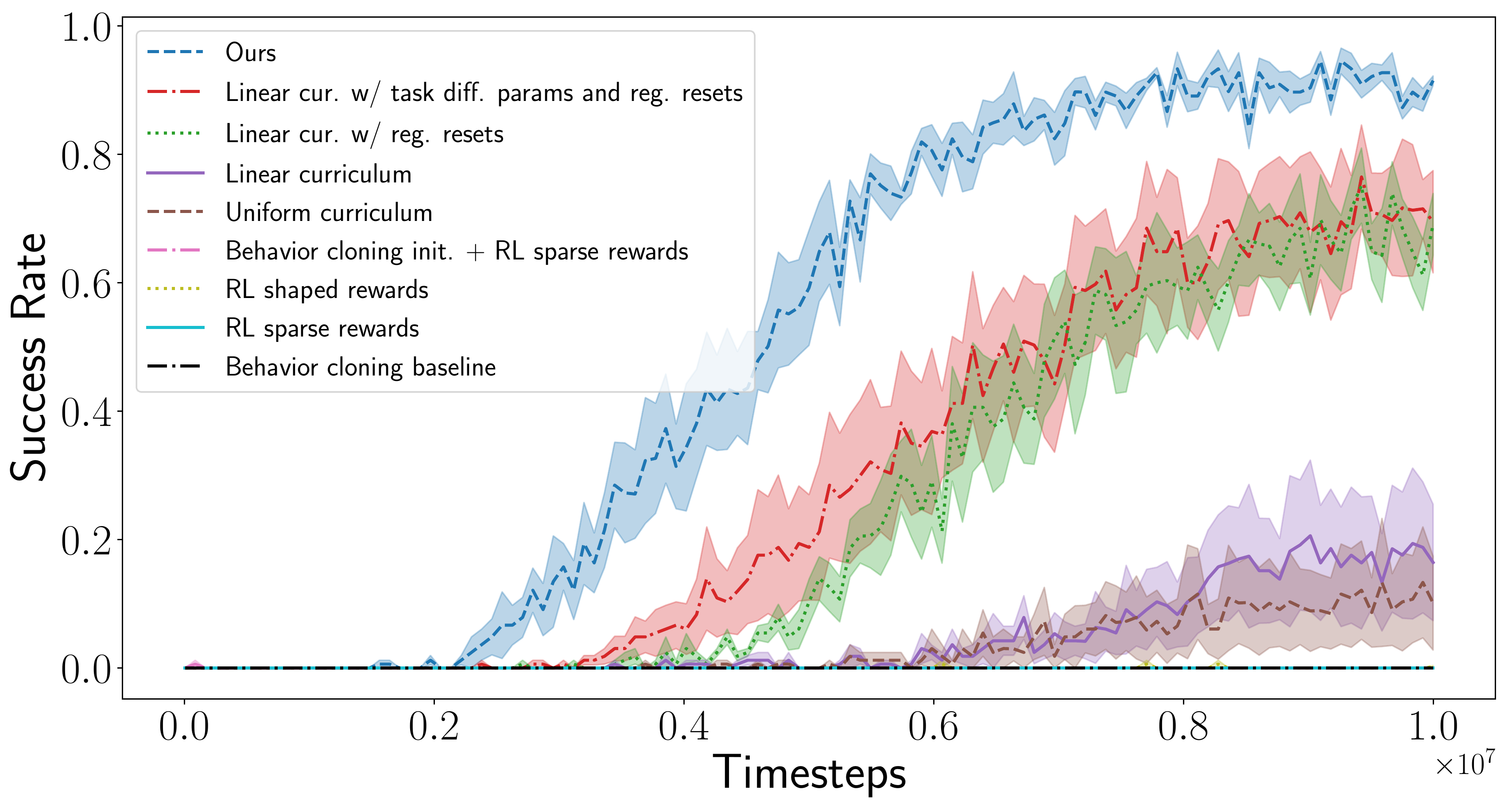}
    \caption{ACGD vs baseline methods for the block stacking task, evaluated in simulation.}
    \label{fig:baselines_stack}
\end{figure}

\begin{figure}[tb]
    \centering
    \includegraphics[width=1.0\columnwidth]{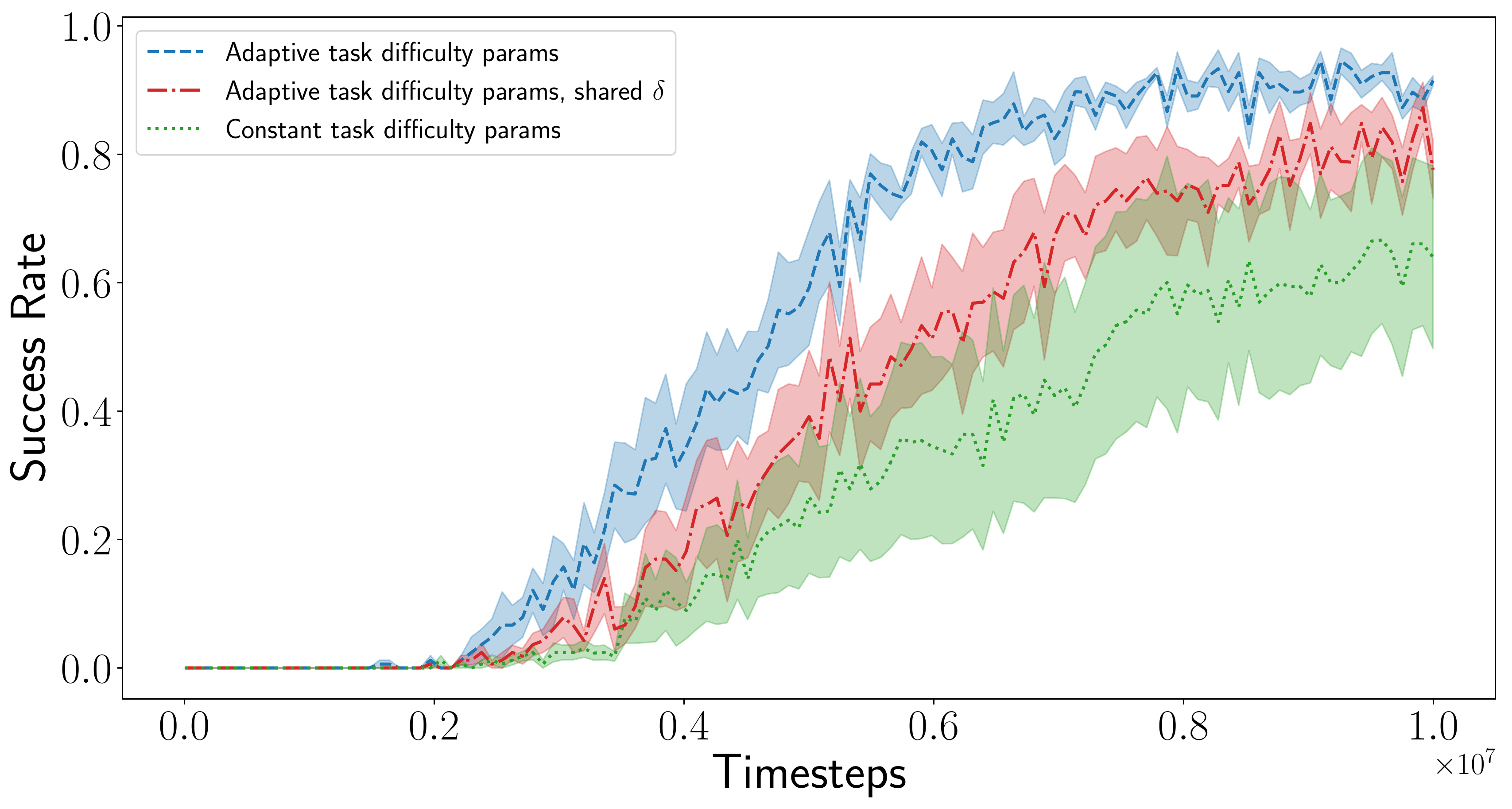}
    \caption{Here we investigate the different methods for choosing the task difficulty for all parameters $\mathcal{H}$ besides the location of demonstration resets. We compare adaptively increasing the difficulty of those parameters during training against always training on the highest difficulty.}
    \label{fig:task_difficulty_stack}
\end{figure}

\begin{figure*}[tb]
    \centering
    \includegraphics[width=0.99\textwidth]{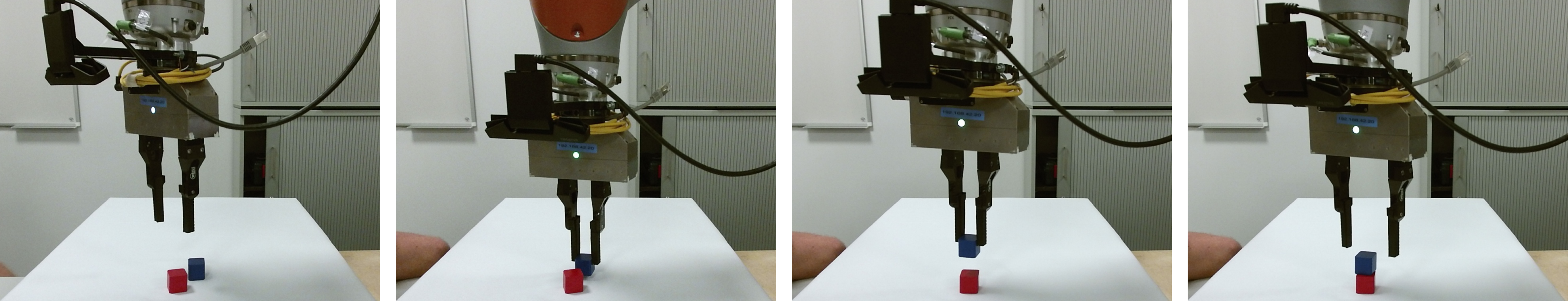}
    \caption{Each image shows the progress of the robot attempting to stack a small blue block on top of a red block. %Note that the policy uses the egocentric images of the in-hand camera and not the third person view shown for visualization.
    }
    \label{coverfigure}
\end{figure*}

\subsection{Experiments in Simulation}\label{exps}
For experiments in simulation we use the PyBullet Physics simulator \cite{Coumans:2015:BPS:2776880.2792704}.
We compare our approach against several baselines: 1) training PPO with sparse and shaped rewards, 2) behavior cloning, 3) PPO with behavior cloning initialization, 4) several standard non-adaptive curriculum learning methods that use demonstrations. 
For both tasks, we recorded a set of 10 manual demonstrations for curriculum learning and an additional set of 100 demonstrations for the BC baseline, both with a 3D mouse.
%The demonstrations do not have to be optimal since the algorithm only uses the states of the recorded trajectories to shape the initial state distribution and discards the actions.
%\footnote{We tried to record succinct demonstrations but it is difficult to manually control several degrees of freedom simultaneously despite using specialized hardware such as a 3D mouse.}

\textbf{Pick-and-stow}. 
The robot has to pick up a small block and place it within a box.
The initial position of the robot and the block randomly change every episode.
A sparse reward of $1-\phi$ is received only after reaching a goal state, where $\phi$ denotes a penalty for collisions of gripper and block with the edges of the box.
The dense reward function for the RL baseline is composed of the Euclidean distance between gripper and block as well as the distance between block and a location above the box plus additional sparse rewards for successfully grasping and placing the block. 

Fig. \ref{fig:comparison_box} shows the task success rates during training.
The results show averages over five different random seeds for every experiment. 
Our method successfully solves the task with an average final success rate of 94\%.
The policy trained with BC achieves a notable success rate of 23\%, but overall lacks consistency having overfitted to the small set of demonstrations.
Interestingly, the policy that used BC as initialization for RL performed poorly and experienced catastrophic forgetting after few timesteps.
Neither RL with sparse rewards nor RL with shaped rewards are able to completely finish the task.
While the former is unable to learn any meaningful behavior, the latter learns to grasp the block but fails to place it inside the box.

\textbf{Block stacking}
%Block stacking is a popular benchmark task for robot manipulation and has been examined by previous work, e.g. in \cite{Deisenroth-RSS-11}, \cite{DBLP:journals/corr/PopovHLHBVLTER17}, \cite{fan2018surreal} and \cite{Zhu-RSS-18}.
To solve the task, the agent has to stack one block on top of the other one, which involves several important subproblems such as reaching, grasping and placing objects precisely. The task features complex, contact-rich interactions that require substantial precision.
We further increased the difficulty of the task compared to prior work by using smaller blocks.
Since it is a task of long episode length that requires at least around 70 steps to be solved with our choice of per-step action ranges, it is particularly well suited for curriculum learning. 

We define a successful stack as one block being on top of the other one for at least 10 timesteps, which indicates stable stacking.
A sparse reward of $1-\phi$ is given after a successful stack where $\phi$ denotes a penalty for the movement of the bottom block during the execution of the task. 
The shaped reward function consists of a mixture of distance functions and sparse rewards similar to the pick-and-stow task.

The training progress and the evolution of difficulty coefficients are shown in Fig. \ref{fig:comparison_stack}.
We see that the success rate is kept in the desired interval through the adaption of the difficulty coefficients until the highest difficulty is reached.
The success rate of evaluation runs (green) is higher due to execution of a deterministic policy.
%We evaluate against the BC and RL baselines, but additionally conduct curriculum learning ablation studies.
Fig. \ref{fig:baselines_stack} shows the results compared to baselines. 
As it is considerably harder than the previous task, none of the curriculum-free baselines are able to solve the stacking task.
In comparison to the uniform and linear reverse curriculum learning variants, our method learns faster and achieves a better final performance, with the final success rate being more than 20\% higher. For the linear curriculum variant start states are sampled linearly further away from the goal during the course of the training.
Our method shows less variance across the seeds, which indicates that the adaptive curriculum improves the stability of learning.
The experiment also demonstrates the importance of not learning exclusively from demonstrations, especially if the amount of demonstration data is limited.
Linear curricula with regular resets, similar to \cite{fan2018surreal}, clearly outperform uniform and linear curriculum learning that are trained only on initial states sampled from demonstrations.

Another advantage of our method is that it can learn substantially more efficient solutions than those provided by the demonstrations. This results from the use of demonstration states, but not the actions taken. For the stacking task, manual demonstration episodes had a mean duration of $164 \pm 17$ transitions, while the learned solution had a mean duration of $73 \pm 15$ transitions. This means that in simulation our trained policy solves the task twice as fast as a human expert operating the robot with a 3D mouse.

We further evaluated how adaptively changing the task parameters $\mathcal{H}$\footnote{not including the trajectory sampling position} for the task difficulty and domain randomization improves the training speed and performance.
In Fig.~\ref{fig:task_difficulty_stack} we compare our full model with the following ablations:
1)~\textit{shared $\delta$} for demonstration resets and regular resets i.e. $\delta_r = \delta_d$, 2) \textit{constant task difficulty}, i.e. adaptive curriculum from demonstrations, but without changing the difficulty of the task parameters $\mathcal{H}$, which were set to the maximum difficulty for the complete training.
%\textit{Constant task difficulty} shows a particularly high variance, some random seeds performed almost as good as the full model, while others do not even achieve $50\%$ success rate.

% for cube stacking we should mention the things that make our version of this task more difficult.
% > small blocks
% > wood blocks i.e. not foam like in google/paper, so more bouncy
% > end of arm camera makes 

%demonstrations: mean 164, std 17
%policy: mean 73, std 15
%Plots for Appendix (maybe):
%1. Shared difficulty vs separate difficulty
%2. Image vs Image+state
%3. Success on real robot vs episode time given

\begin{table}[t]
    \centering
    \begin{tabular}{c c c c c}
	  	Task &   Train & Test &  Trials & Success Rate \\
	  	\hline
   		\hline
	  	Block stacking  			& \multirow{2}{*}{Simulation}  & \multirow{2}{*}{Simulation} & 91/100 & 91\%   \\
	  	Pick-and-stow               &   &  & 94/100 & 94\% \\
   		\hline
	  	Block stacking  			& \multirow{2}{*}{Simulation}  & \multirow{2}{*}{Real} & 12/20 & 60\%   \\
	  	Pick-and-stow               &  &  & 17/20 & 85\% \\
	  	\hline
    \end{tabular}
    \caption{Success rates in simulation and real-world.}
    \label{tab:success}
    \vspace{-2mm}
\end{table}

\begin{table}[t]
\centering
\begin{tabular}{|l|}
\hline
\textbf{Difficulty Parameters:} $\mathcal{H}_d$\\
\hline
\begin{minipage}[t]{.94\columnwidth}%
bounciness of objects,
num. steps stacked  for task success, 
gripper speed, 
position offset of relative Cartesian position control, 
camera field of view,
camera position and orientation,
block color,
table color,
camera image brightness,
camera image contrast,
camera image hue,
camera image saturation,
camera image sharpness,
camera image blur,
light direction
\end{minipage}\\
\\
\textbf{Additional Regular Reset Parameters:} $\mathcal{H}_a$ \\
\hline
\begin{minipage}[t]{.94\columnwidth}%
initial gripper height, 
lateral gripper offset, 
distance between blocks, 
min. final block vel.   for task success,
table height,
height of the robot base, 
initial gripper rotation,
block size
\end{minipage}\\
\hline
\end{tabular}
\caption[Task Parameters]{Task parameters used to adapt the difficulty of the stacking task. When initializing from demonstration states only a subset $\mathcal{H}_d$ can be randomized, regular resets randomize $\mathcal{H}_r = \mathcal{H}_d \cup \mathcal{H}_a$.} \vspace{-1em}
\label{tab:task_params}
\end{table}

\subsection{Experiments with Real Robot}
%Next we evaluate our policies, trained for real tasks with domain randomization, on the real robot.
We applied the trained policies without any additional fine-tuning on the real robot. Our results are shown in Table~\ref{tab:success}.
We see that despite incurring a performance penalty by evaluating on the real robot, the policies transfer with a good success rate of 85\% for pick-and-stow and 60\% for block stacking. Both task were evaluated with a constant episode length of 300 timesteps (15 seconds).
Within the given time frame, the policy can attempt a second trial after a failed first execution of the task.
This shows the advantages of a policy learned in closed-loop as it implicitly aims to re-grasp the block in case of a failed stacking or stowing attempt.
Using small wooden blocks with an edge length of only 2.5cm requires a highly precise actuation.
as the block tends to fall over if not placed precisely or dropped from a too large height. In contrast, polices learned with non-adaptive curriculum learning baselines were unable to achieve the sim-to-real transfer.
It is difficult to compare performance with previous approaches due to the lack of clear benchmark setups.
In a related approach, Zhu \etal~\cite{Zhu-RSS-18} performed zero-shot sim-to-real transfer of a block stacking task, however, they use large deformable foam blocks for stacking. These are easier to grasp because they are made of foam and easier to stack because they are larger and have more friction.
They report a success rate of 35\% for stacking on the real robot, which is lower than ours.

% \argus{what are the modalities of failure?}
% \lukas{eg. blocks dissappear from camera image after unsuccessful grasping attempt, time is up,  ...}

\section{CONCLUSIONS}

The proposed Adaptive Curriculum Generation from Demonstration (ACGD) method enables robust training of policies for difficult multi-step tasks. It does this by adaptively setting the appropriate task difficulty for the learner by controlling where to sample from the demonstration trajectories and which set of simulation parameters to use. This unified treatment of demonstration sampling and domain randomization as task difficulty improves training. In combination with domain randomization the method can train policies in simulation that achieve good success rates when evaluated on a real robot.
% \thomas{It would be nice to pick up on the questions from the beginning of the experimental section and answer them in the conclusions.}

\FloatBarrier
\newpage

% \section*{ACKNOWLEDGMENT}
% We would like to thank the Freiburg Graduate School of Robotics, the  BrainLinks-BrainTools Cluster of Excellence (DFG EXC 1086), the Intel Network on Intelligent Systems, and the EU Horizon 2020 project Trimbot2020 for supporting this project.

%%%%%%%%%%%%%%%%%%%%%%%%%%%%%%%%%%%%%%%%%%%%%%%%%%%%%%%%%%%%%%%%%%%%%%%%%%%%%%%%

\bibliographystyle{IEEEtran}
\bibliography{references}
%\bibliography{IEEEabrv,IEEEexample}

\clearpage
\newpage
\section*{APPENDIX}
% \subsection{Control Details}
% This setup consists of a KUKA iiwa 14 r820 with a WSG-50 gripper to which 3D printed finger-tips have been attached.
% A 3D printed mount is also used to attach a Intel RealSense SR300 RGB-D camera of which we use only the RGB image.
% We control the robot in Cartesian coordinate space with the control reduced to 4 continuous spatial DoF, $(x, y, z, \text{rotation})$.
% There is an additional DoF of the gripper opening, which is controlled as a continuous DoF and acts as discrete action triggered when a threshold is passed.
% The robot is controlled at a rate of 20Hz.
% The controls are interpreted as being in the cameras coordinates system and are converted back into world coordinates using robot state information.
% The camera is aligned so that the camera can see the finger tips of the gripper, this allows running policy networks without memory.
% A good alignment is obtained by calibrating the simulation to align with images obtained from the real system.
% This is done by optimizing IoU between segmented gripper-tips as measured by IoU of the chroma-key segmentations of finger tips.
% Similarly simulation dynamics are also calibrated to observed responsiveness to actions. 

\subsection{Hyperparameters}
Table \ref{tab_app_param} shows a list of hyperparameters that were used in the experiments.
\begin{table}[h!]
\centering
\begin{tabular}{c c}
\textbf{Hyperparameter} & \textbf{Value} \\
\hline
\hline
Adam learning rate & $2.5\times10^{-4}$ \\
Adam $\epsilon$ & $1 \times 10^{-5}$ \\ 
Discount $\gamma$ & 0.99 \\
GAE $\tau$ & 0.95 \\
Entropy coefficient & 0.01 \\
Value loss coefficient & 0.5 \\
Max grad. norm & 0.5 \\
Number of actors & 8 \\
Minibatch size & $8 \times 512$ \\
Num. epochs & 4 \\
Clip param. & 0.1 \\
Training steps & $10^7$ \\
Interval $[\alpha, \beta]$ & $[0.4, 0.6]$ \\
Increment $\epsilon$ & 0.002 \\
Number of demonstrations & 10
\end{tabular}
\caption[Hyperparameters]{Default hyperparameters that were used in the experiments unless stated otherwise.}
\label{tab_app_param}
\end{table}

\subsection{Additional Experiments}
\begin{figure}[h!]
    \centering
    \includegraphics[width=1.0\columnwidth]{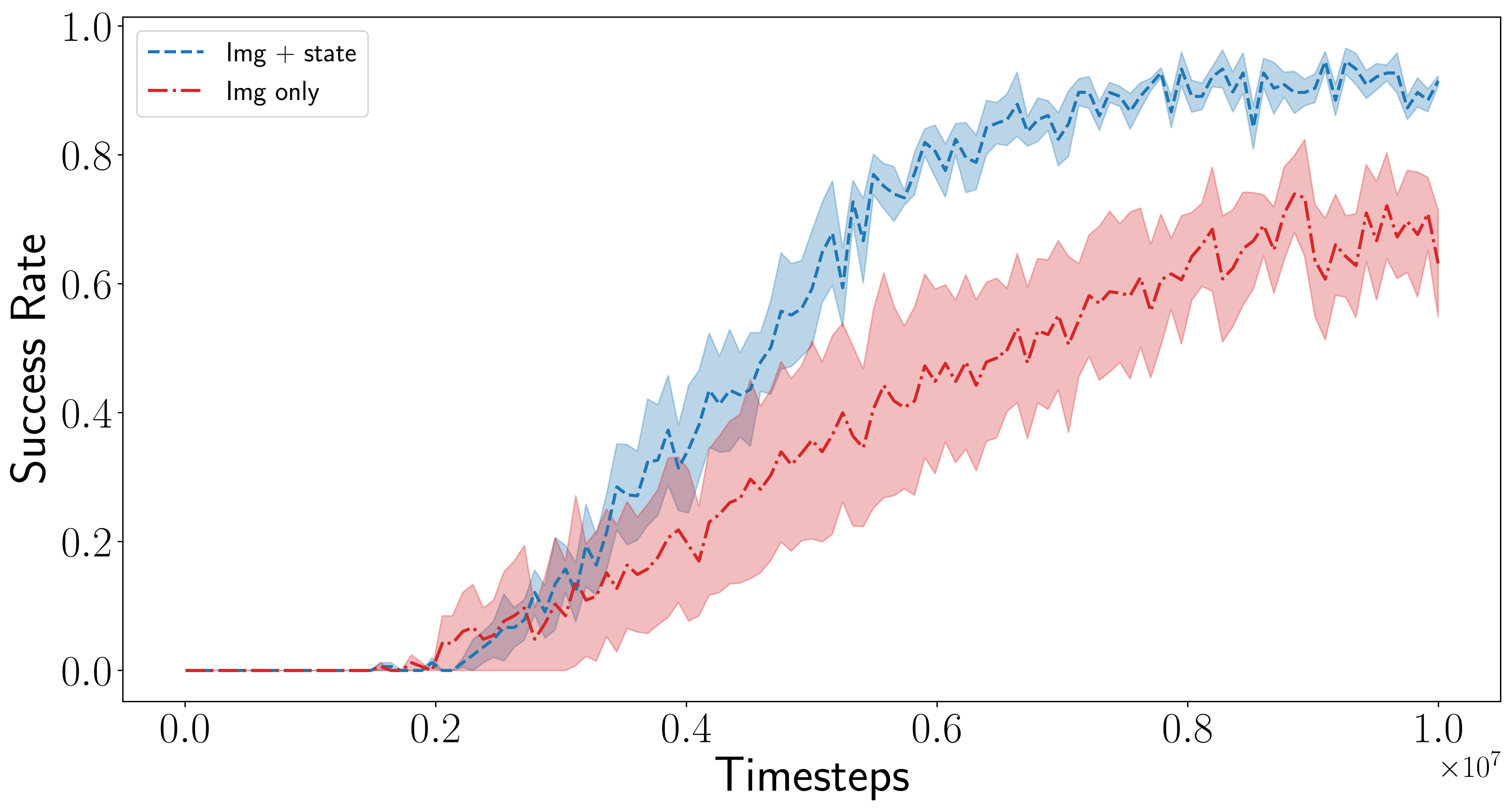}
    \caption{Observation experiment. We evaluate how the performance depends on the input modality.
  \textit{Img only} learns purely from the camera images.  We reduced the network architecture to the CNN part of the default network.
  The full model clearly outperforms the \textit{Img only} network.
  We assume, that this is because the state vector contains valuable information that is not or only insufficiently contained in the images.
  The plot shows averages and standard deviations of the evaluation success rate over five runs with different random seeds.}
    \label{fig:input}
    
\end{figure}

\begin{figure}[h!]
    \centering
    \includegraphics[width=1.0\columnwidth]{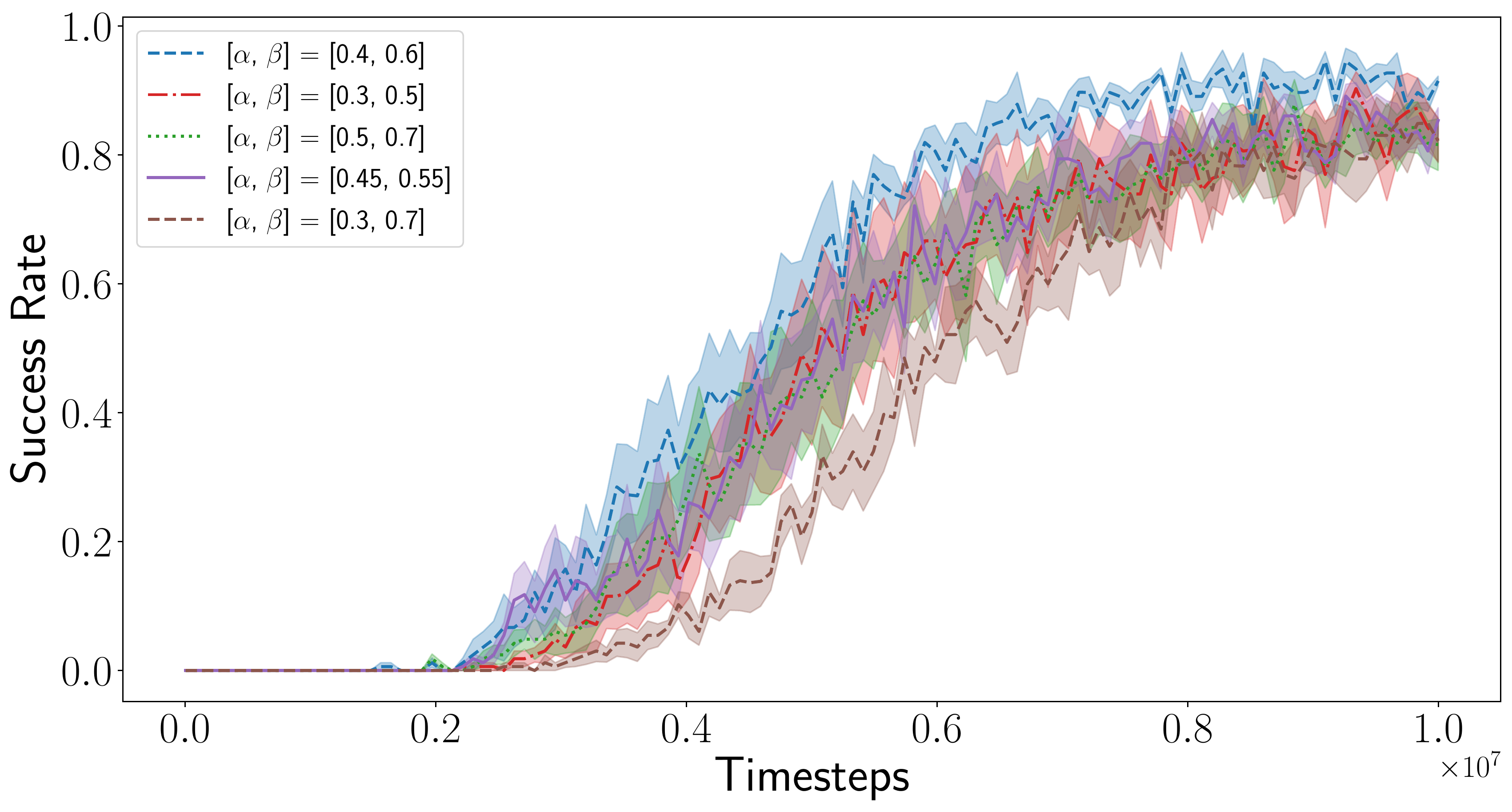}
    \caption{Reward Interval Experiment. In this experiment, we evaluate the influence of difference values for $[\alpha, \beta]$.
  The interval of $[0.4, 0.6]$ shows the best performance and is used in the paper, but the other runs also achieve good scores.
  The plot shows averages and standard deviations of the evaluation success rate over five runs with different random seeds.}
    \label{fig:interval}
    \vspace{-2em}
\end{figure}

\begin{figure}[h!]
    \centering
    \includegraphics[width=1.0\columnwidth]{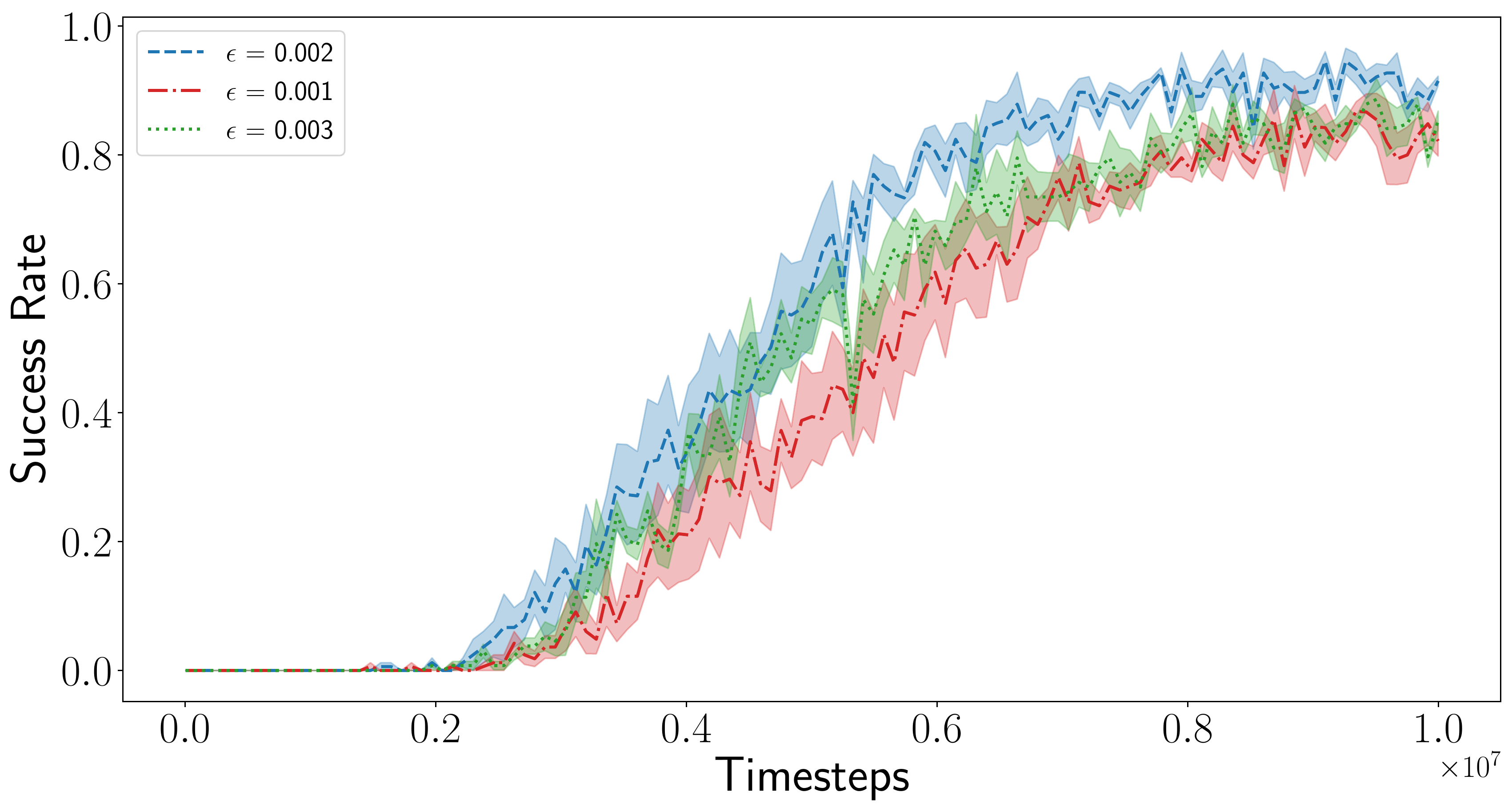}
    \caption{Difficulty Increment Experiment. Here, we compare how different values for $\epsilon$ influence the training speed and final performance.
  All three values show similar training curves with our default value of $\epsilon=0.002$ being slightly better than the rest.
  The plot shows averages and standard deviations of the evaluation success rate over five runs with different random seeds.}
    \label{fig:increment}
\vspace{20em}
\end{figure}

Additional experiments were conducted for the block stacking task described in Section \ref{exps}.
We investigate the choice of input modalities (Fig. \ref{fig:input}) and the impact of different values for the hyperparameters reward interval $[\alpha, \beta]$ (Fig. \ref{fig:interval}) and difficulty increment $\epsilon$ (Fig. \ref{fig:increment}).
\vspace{2em}
\FloatBarrier

% \subsection{Reward Functions}
% \textbf{Pick-and-stow}
% \textbf{Block stacking}
% The sparse reward function consists of a penalty $\phi$ for the movement of the bottom block during the execution of the task.
% It is defined as:
% \begin{equation}
%     \phi = \text{tanh}(\text{cos}^{-1}(2\langle \Vec{q}_{start}, \Vec{q}_{end} \rangle^2 - 1) + 10 * \| \Vec{pos}_{start} - \Vec{pos}_{end}\|),  
% \end{equation}
% where $\Vec{pos}$ and $\Vec{q}$ denote position and orientation (expressed as quaternion) of the bottom block.

\end{document}